\documentclass[sn-mathphys]{sn-jnl}

\jyear{2021}%

\theoremstyle{thmstyleone}%
\theoremstyle{thmstyletwo}%

\theoremstyle{thmstylethree}%
\usepackage{subfigure}

\usepackage{flushend}
\usepackage{hyperref}
\usepackage[export]{adjustbox}
\usepackage{multirow}
\newcommand{\tabincell}[2]{\begin{tabular}{@{}#1@{}}#2\end{tabular}}
\usepackage{verbatim}

\begin{document}

\title[2s-CNN for Musculoskeletal and Neurological Disorders Prediction]{A Two-stream Convolutional Network for Musculoskeletal and Neurological Disorders Prediction}


\author[1]{\fnm{Manli} \sur{Zhu}}\email{manli.zhu@northumbria.ac.uk}

\author[2]{\fnm{Qianhui} \sur{Men}}\email{qianhui.men@eng.ox.ac.uk}

\author[3]{\fnm{Edmond S. L.} \sur{Ho}}\email{Shu-Lim.Ho@glasgow.ac.uk}

\author[4]{\fnm{Howard} \sur{Leung}}\email{howard@cityu.edu.hk}

\author*[5]{\fnm{Hubert P. H.} \sur{Shum}} \email{hubert.shum@durham.ac.uk}

\affil[1]{\orgdiv{Department of Computer and Information Sciences}, \orgname{Northumbria University}, \orgaddress{ \city{Newcastle upon Tyne}, \country{UK}}}

\affil[2]{\orgdiv{Department of Engineering Science}, \orgname{University of Oxford}, \orgaddress{\city{Oxford}, \country{UK}}}

\affil[3]{\orgdiv{School of Computing Science}, \orgname{University of Glasgow}, \orgaddress{ \city{Glasgow}, \country{UK}}}

\affil[4]{\orgdiv{Department of Computer Science}, \orgname{City University of Hong Kong}, \orgaddress{\city{Kowloon}, \country{Hong Kong}}}

\affil*[5]{\orgdiv{Department of Computer Science}, \orgname{Durham University}, \orgaddress{\city{Durham}, \country{UK}}}


\abstract{Musculoskeletal and neurological disorders are the most common causes of walking problems among older people, and they often lead to diminished quality of life. Analyzing walking motion data manually requires trained professionals and the evaluations may not always be objective. To facilitate early diagnosis, recent deep learning-based methods have shown promising results for automated analysis, which can discover patterns that have not been found in traditional machine learning methods. We observe that existing work mostly applies deep learning on individual joint features such as the time series of joint positions. Due to the challenge of discovering inter-joint features such as the distance between feet (i.e. the stride width) from generally smaller-scale medical datasets, these methods usually perform sub-optimally. 
As a result, we propose a solution that explicitly takes both individual joint features and inter-joint features as input, relieving the system from the need of discovering more complicated features from small data. Due to the distinctive nature of the two types of features, we introduce a two-stream framework, with one stream learning from the time series of joint position and the other from the time series of relative joint displacement. 
We further develop a mid-layer fusion module to combine the discovered patterns in these two streams for diagnosis, which results in a complementary representation of the data for better prediction performance.
We validate our system with a benchmark dataset of 3D skeleton motion that involves 45 patients with musculoskeletal and neurological disorders, and achieve a prediction accuracy of 95.56\%, outperforming state-of-the-art methods.
}




\maketitle

\section{Introduction}
\label{sec:introduction}

Musculoskeletal and neurological disorders, such as joint problems, muscle weaknesses, and neurological defects (Table \ref{example}), are the most common causes of walking problems among older people, and they often lead to diminished quality of life. The prevalence of gait and balance abnormalities appears more than 60\% in people aged over 80 years \cite{AgeBackground}. Gait analysis is a popular method for diagnosing these disorders. However, analyzing walking data manually requires trained professionals, and the evaluations may not always be objective \cite{traditionalgait2014}. We focus on proposing a low-cost automated tool for the early prediction and effective therapy monitoring of musculoskeletal and neurological disorders. First, it allows early intervention clinical care before the disorders develop into bigger health issues. Second, it supports clinicians make a more robust diagnosis by providing a computer-aided indicator and helps them effectively in monitoring patients' health conditions.


\begin{table}[htbp]
	\centering
	\caption{Class of disorders and examples}
	\label{example}
	\setlength{\tabcolsep}{5pt}
	\begin{tabular}{cccc}\hline
		Class & Joint Problem & Muscle Weakness & Neurological Defect\\ \hline
		
		\multirow{3}*{Examples} & Sprains \cite{sprains} &Muscular dystrophy \cite{muscular-dystrophy}& Epilepsy \cite{epilepsy} \\
		
		& Tendinitis \cite{tendinitis} &Spinal muscular atrophy \cite{spinal-muscular-atrophy}& Alzheimer's disease \cite{jad2020}\\
		
		& Osteoarthritis \cite{osteoarthritis} &Muscle fatigue \cite{muscle-fatigue}& Parkinson's disease \cite{Xiapd2020} \\\hline
		
	\end{tabular}
\end{table}


Machine learning (ML) and deep learning (DL) have been widely used for automatically identifying health issues \cite{RNN2019,PD2019,PD2020,mergeSVMPD2020,gaitalter2019,McCay:TNSRE2022}. For example, by using 3D motion analysis, support vector machines (SVM) were applied for Parkinson’s disease classification in \cite{gaitML} from gait signals. Begg et al. \cite{SVMJointAngles} also classified young-old gait types with SVM from joint angle features. However, in the medical domain, such conventional approaches have restricted ability to model complicated data due to their limited capacity. They require considerable understanding and expertise for feature representation, i.e., feature engineering, since they have limited capability in processing raw data \cite{mlVSdl2018}. While deep learning allow multi-level abstractions of the raw data for decision making due to its deep architecture of non-linear hidden layers \cite{dloverview}. It facilitates the automatic diagnosis of disorders. While DL approaches are more advantageous with their deep hidden layer architectures. For instance, Davarzani et al. \cite{3Dgait1} used the long short-term memory (LSTM) network for human gait recognition from foot angle movements and achieved better performance than linear regression. McCay et al. \cite{KevinIEEEAccess} applied deep convolutional networks and achieved better prediction performance in cerebral palsy diagnosis than SVM, decision tree, and k-nearest neighbors algorithms (KNN).

Among different DL architectures, convolutional neural networks are very popular and have achieved promising performance on many diagnostic tasks \cite{AD-JMS,depression-JMS,emotion-JMS}. However, these networks heavily rely on large datasets to avoid overfitting \cite{jbd-overfitting}. Unfortunately, large datasets are often not available in medical video/image analysis due to the restrictions on sharing data publicly in this domain. This makes it difficult to discover inter-joint correlations that are important for capturing coordination among different joints of human gait from raw joint features. We also observed that the majority of existing work only applies DL on individual joint features such as the time series of joint positions \cite{KevinIEEEAccess,BHI-Manli}. As a consequence, these methods usually perform sub-optimally on smaller-scale medical datasets. 

In this paper, we propose a two-stream CNN (2s-CNN) framework that explicitly takes both individual joint features and inter-joint features as input, allowing more effective discovery of features for disorder prediction from small data. The two different sets of features reflect different patterns of human motion data, i.e., the joint position describes the geometric location of an individual joint, and the relative joint displacement extracts the correlations of inter joints. To analyze the coordination and synchronization of different body part movements for better modelling walking motion, it is important to extract features from different joints simultaneously \cite{McCay:TNSRE2022}. As such, we include the relative joint displacement feature which explicitly contains joint coordination patterns to guide the DL model to discover the essential inter-joint correlations for better diagnosis outcomes. To optimally model such distinctive feature, our network consists of two separate streams - a 3D joint position stream (3DJP-CNN) learning from the time series of joint position, and a 3D relative joint displacement stream (3DRJDP-CNN) learning from the time series of relative joint displacement. We further introduce a mid-layer fusion module fusing two single streams, which facilitates capturing both the individual joint information and inter-joint correlations, resulting in a more complementary representation of the data. 

An extensive evaluation of the proposed framework is performed on the 3D skeletal motion dataset \cite{Worasak2018}. Different from \cite{Worasak2018}, which evaluated the performance of a single type of feature on different off-the-shelf ML-based classifiers, our DL framework takes two types of features as input to model different aspects of the skeletal motion to achieve state-of-the-art performance. We further report the per-class classification performance to show the feasibility of deploying the framework as a patient diversion system, rather than only the overall average classification accuracy as in \cite{Worasak2018}. We justify our two-stream framework design by demonstrating its superior performance to the individual streams' as a baseline study. To stimulate the research in this area, we open-source this project by releasing the source code for further validation and development. Our processed dataset features a standardized format, the evaluation protocol, as well as the augmented data for minimizing data bias. They can be downloaded at \href{https://github.com/zhumanli/2s-CNN}{https://github.com/zhumanli/2s-CNN}.
	
This paper is organized as follows. Data preparation is given in section \ref{sec:DataPreparation}. Section \ref{sec:TwoStream} presents the methodology. Experimental results and ablation studies are provided in section \ref{sec:ExperimentalResults}. Section \ref{sec:Conclusion} concludes the research.

\section{Data Preparation}
\label{sec:DataPreparation}
Here, we explain the public benchmark dataset that we employ in this research (Section \ref{sec:dataset}), and our data augmentation strategy to deal with the data bias problem commonly found in medical datasets (Section \ref{sec:DataAug}).

\subsection{The Dataset}\label{sec:dataset}
We employ the dataset created by Rueangsirarak et al. \cite{Worasak2018}. It consists of 4 classes and 45 walking motions, i.e., 10 healthy, 4 joint problems, 18 muscle weakness, and 13 neurological defect motions. They were performed by 45 subjects, who were aged between 61 and 91 years old. The subjects were diagnosed to be one of the 4 classes by three medical doctors. The standard clinical test was used by medical experts for voluntary applicants' screening and approval, e.g., applicants could walk without any assistance and had no other medical disorder history that could affect walking, and the details can be found in \cite{Worasak2018}. By applying a randomly sampled and population-based study, 5 male subjects and 40 female subjects were selected from the applicants' approved list. The gender bias reflects the bias of the voluntary applicants in such a community.

\begin{figure*}[htbp]
	\centering
	\subfigure[]{
		\begin{minipage}{0.45\linewidth}
			\centering
			\includegraphics[scale=0.7]{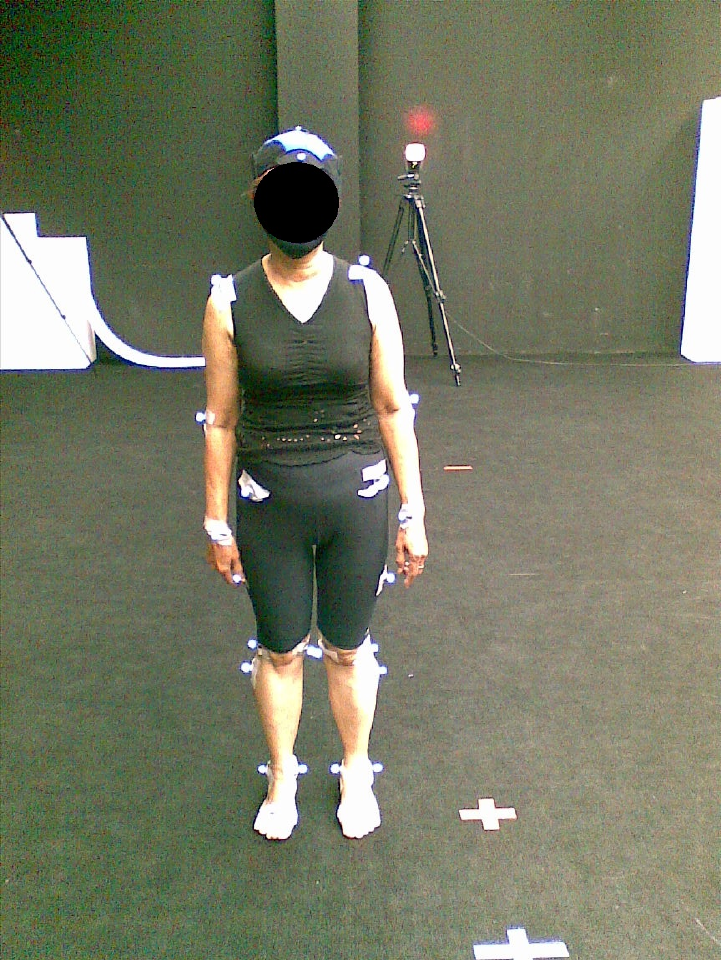}
		\end{minipage}
	}
    \quad
	\subfigure[]{ 
		\begin{minipage}{0.45\linewidth}
			\centering	\includegraphics[scale=0.7]{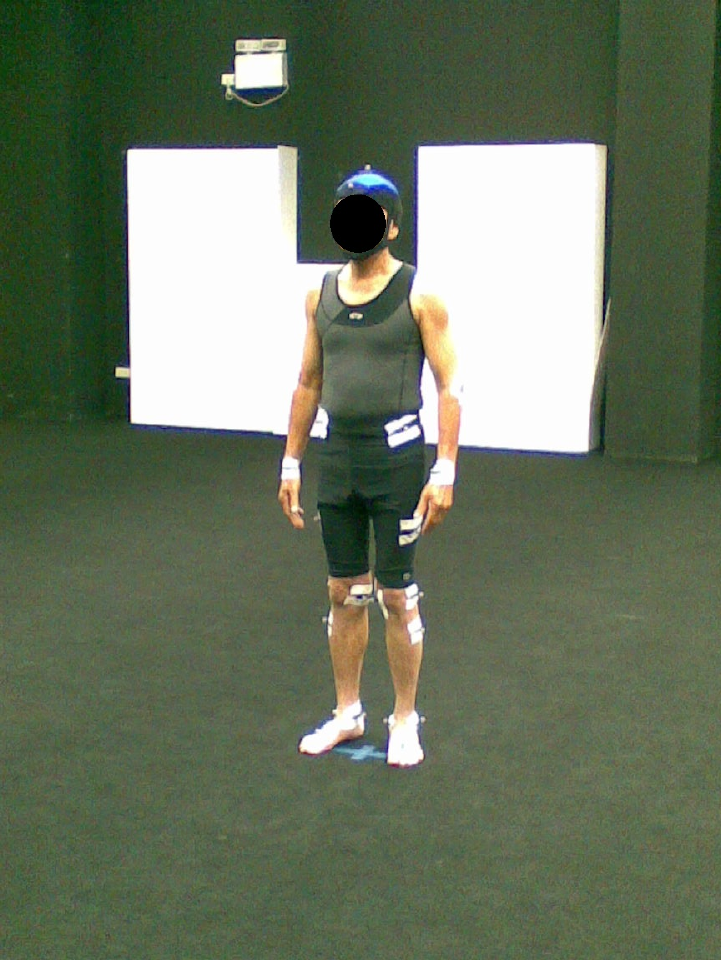} 
		\end{minipage}
	}
	\caption{The optical motion capture system used with the Helen Hayes marker set structure.} 
	\label{patient} 
\end{figure*}

The data were captured using the Motion Analysis$^\circledR$ optical motion capture system \cite{motioncapture} with fourteen Raptor-E optoelectronic cameras sampling at 100 Hz. The subjects were required to wear a motion capture suit, attached with a markers set on their body based on the Helen Hayes marker set structure \cite{helenmarker}, as shown in Fig \ref{patient}. They were asked to walk naturally at their normal walking speed for 10 meters. The output of the motion capture system is 3D markers' time-series positions. This optical marker-based capture method is advantageous because the captured data are more accurate than the markerless method, thereby facilitating more accurate diagnosis. Finally, the 3D positions of the joints are estimated from the marker locations by fitting a virtual character with similar body proportions to each subject in the software Autodesk MotionBuilder.

\begin{figure*}[htbp]
	\setlength{\abovecaptionskip}{5pt} 
	\centering
	\includegraphics[width=0.7\linewidth]{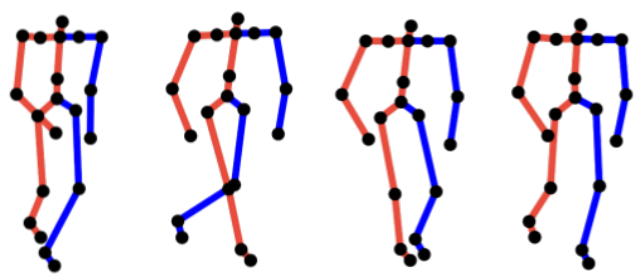}
	\caption{\footnotesize{The sample of a human walking cycle} (progressing from left to right)}
	\label{Fig: walkcycle}
\end{figure*}

\begin{figure}[htbp]
    \setlength{\abovecaptionskip}{2pt} 
    \hspace{0.4in} 
	\begin{minipage}[c]{0.4\linewidth}
		\includegraphics[scale=0.18]{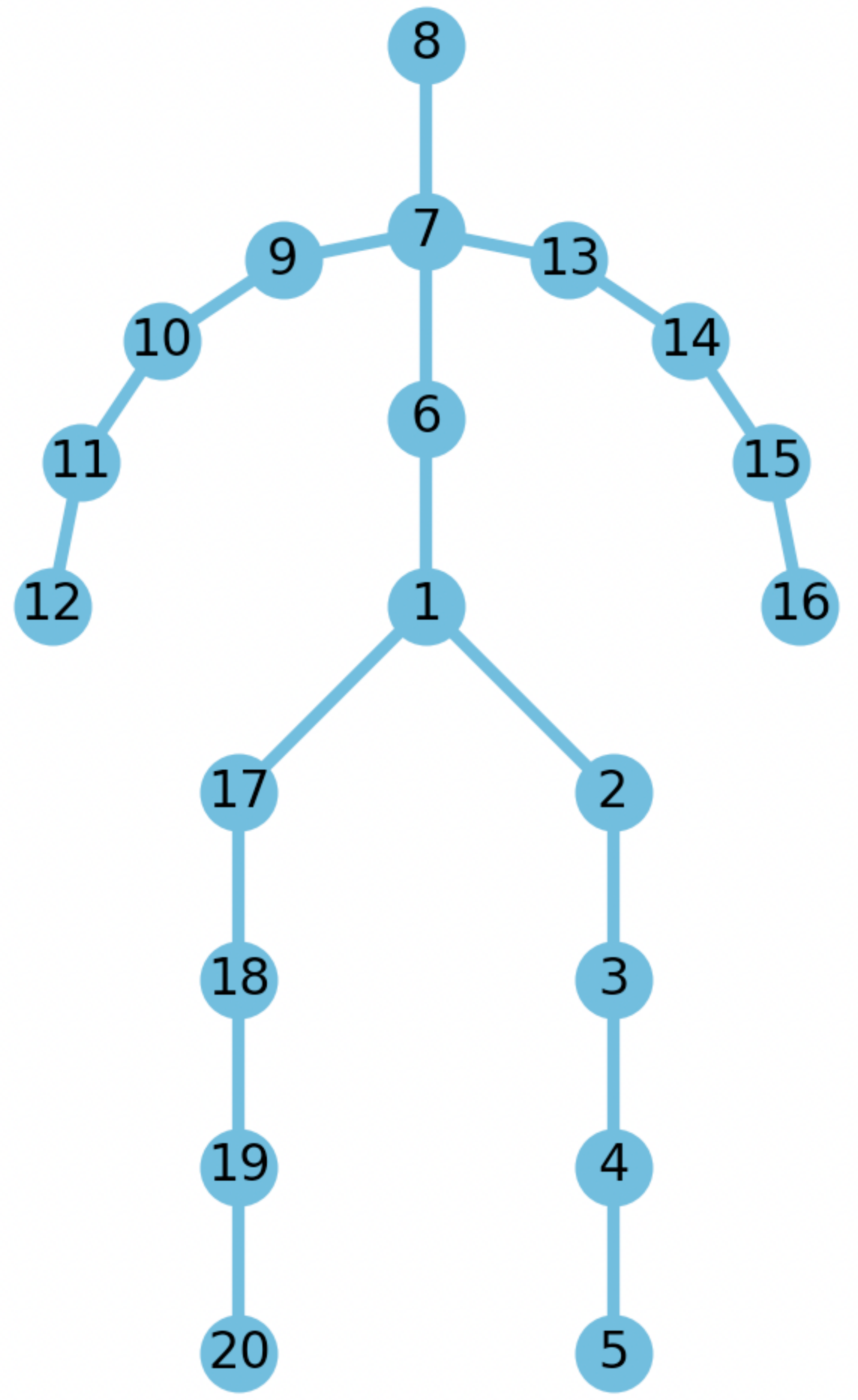}
	\end{minipage}
	\hspace{0.1in} 
	\begin{minipage}[c]{0.4\linewidth}
        \footnotesize
		1: Hips \\
		2: Right Upper Leg \\
		3: Right Leg \\
		4: Right Foot \\
		5: Right Toes \\
		6: Spine \\
		7: Neck \\
		8: Head \\
		9: Left Shoulder \\
		10: Left Arm \\
		11: Left Forearm \\
		12: Left Hand \\
		13: Right Shoulder \\
		14: Right Arm \\
		15: Right Forearm \\
		16: Right Hand \\
		17: Left Upper Leg \\
		18: Left Leg \\
		19: Left Foot \\
		20: Left Toes \\
	\end{minipage}
	\caption{\footnotesize{The overview of the skeleton structure}}
	\label{Fig: skeleton}
\end{figure}

We perform normalization in both temporal and spatial domains using Autodesk MotionBuilder. For the temporal dimension, we extract three entire walking cycles of each subject from the raw data, and the intermediate cycle (includes the complete stance and swing phases) among three walking cycles is kept since it better represents the subjects' normal walking motions. 
As the duration of each walking cycle is different, we temporally scale them using the linear interpolation \cite{linearinterpolation2002} such that all motions will have the same duration. Fig. \ref{Fig: walkcycle} illustrates a sample of a walking cycle. For the spatial dimension, the walking motion is normalized using rotation and translation operations such that the start positions and moving directions are the same among all motions and are comparable. Notice that end effectors such as end toes were removed as they are less informative and noisy. We then extracted 20 main joints that formed the final skeleton structure as shown in Fig. \ref{Fig: skeleton}. 

A major advantage of this dataset is that it includes multiple disorders, it facilitates the patient diversion system. Many existing computer-aided diagnostic systems of musculoskeletal and neurological disorders are limited to binary classification \cite{binaryc1,binaryc2}, i.e., they simply differentiate between healthy and unhealthy data without specific types of disorders for unhealthy patients. With a model developed based on this multi-class dataset, patients can be transferred to a specific department as early as possible thus human resources can be greatly reduced.
	
\subsection{Data Augmentation} \label{sec:DataAug}
The used dataset in this study is challenging due to its multiple, small-scale, and biased classes of disorders. The number of training samples is a critical influencing factor on the generalization ability of DL models. To make the DL model generalize well, augmentation techniques such as random scaling, noise addition, sign inversion, and motion reverse were applied to generate more training samples in cerebral palsy prediction \cite{DimitriosAccess}. However, the aforementioned techniques only focus on intra-class variations. As a result, the augmented data may not be effective in alleviating the inter-class similarity problem as multiple class labels have to be considered. Here, we apply the synthetic data augmentation method \textit{mixup} \cite{mixup}, resulting in an unbiased, 4 times larger dataset. This size is reasonable for our model learning, which prevents overfitting and generates good performance from our experiments.

\textit{Mixup} \cite{mixup} fits better to our dataset compared to data agumentation methods such as \textit{Gaussian noise} \cite{rehabnet2019} and \textit{SMOTE} \cite{smote2002}. It generates synthetic data from real samples of different classes in an interpolation manner rather than adding random noise or generating within the same class. This facilitates the generation of reasonable synthetic data in scenarios when the data have specific structures (e.g., the hierarchical structure for body joint positions and angles on a human body) and limited samples are available.

The data augmentation operation is described as follows:
\begin{equation}
X_{v}=\lambda X_{a}+(1-\lambda) X_{b}
\end{equation}
where $X_{v}$ is a synthetic sample, $X_{a}$ and $X_{b}$ are two samples randomly selected from class $a$ and $b$ respectively, and $\lambda \in[0,1]$ represents how much contribution to the synthetic data from the two original ones.
We empirically set $\lambda=0.9$ based on the performance of our two-stream CNN framework, and the same label as $X_{a}$ is assigned to the generated sample $X_{v}$. By this, small inter-class similarities will be introduced which encourages the neural network to learn more discriminative deep representations to differentiate samples from different classes.

The augmentation is done for different cross-validations. Concretely, for each cross-validation, one fold is used for testing, and the remaining folds are firstly used to generate synthetic samples, then together with the generated synthetic samples are used for training. Besides presenting the number of samples of the original data, the statistic of data with augmentation is also given in Table \ref{datastic}.
	
\begin{table}[htbp]
	\centering
	\caption{The statistics of the original data and the augmented data}
	\label{datastic}
	\setlength{\tabcolsep}{6pt}
	\begin{tabular}{c c c}\hline
		Class & Original data & With data augmentation \\ \hline
		Healthy & 10 & 45 \\
		Joint Problem & 4 & 45 \\
		Muscle Weakness & 18 & 45 \\
		Neurological Defect & 13 & 45 \\ \hline
		Overall & 45 & 180 \\\hline
	\end{tabular}
\end{table}

\section{Methodology}
\label{sec:TwoStream}
\begin{figure*}[htbp]
	\setlength{\abovecaptionskip}{5pt} 
	\centering
	\includegraphics[width=1\linewidth]{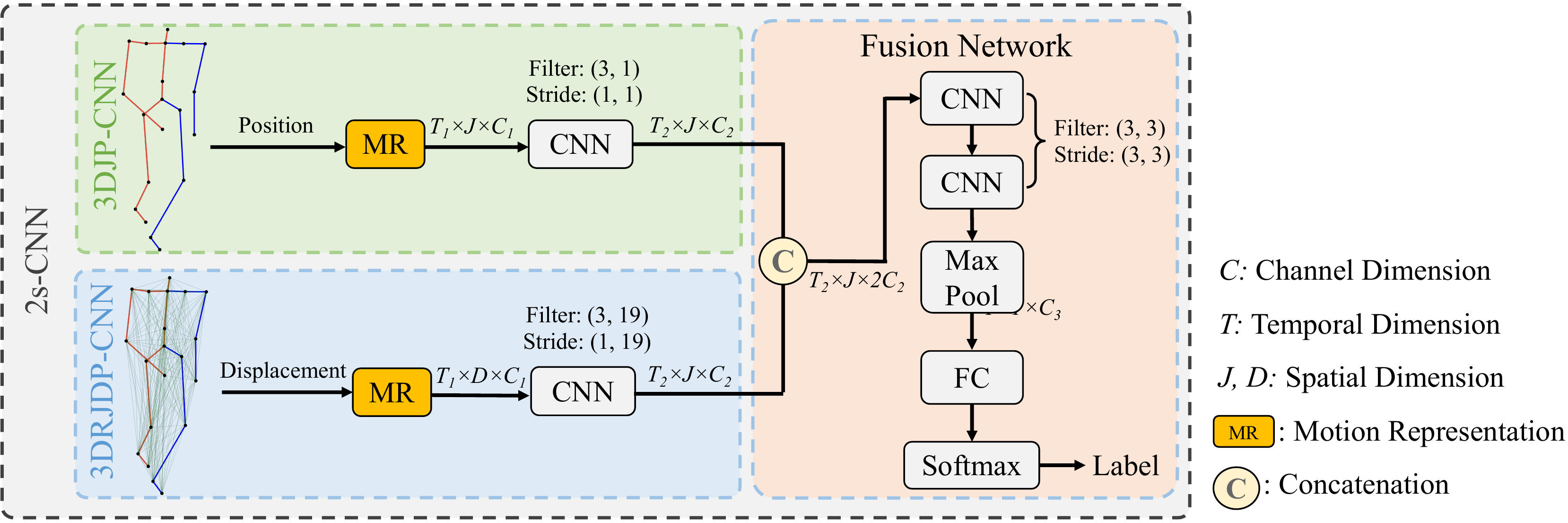}
	\caption{\footnotesize{Overview of our proposed two-stream framework}}
	\label{Fig: framework}
\end{figure*}

We propose a two-stream convolutional neural network that explicitly takes two types of features as input, Fig. \ref{Fig: framework} illustrates its architecture. The 3DJP-CNN stream (Section \ref{sec: JPStream}) aims at modelling individual joint time series from joint positions, and the 3DRJDP-CNN stream (Section \ref{sec: RJDPStream}) aims at extracting inter-joint correlations from relative joint displacements. The mid-layer fusion module (Section \ref{sec: fusion}) fuses the two single streams' high-level output features to take advantage of both individual joint information and inter-joint correlations.

\subsection{The Joint Position Stream}
\label{sec: JPStream}
\begin{figure}[htbp]
	\setlength{\abovecaptionskip}{5pt} 
	\centering
	\includegraphics[scale=0.5]{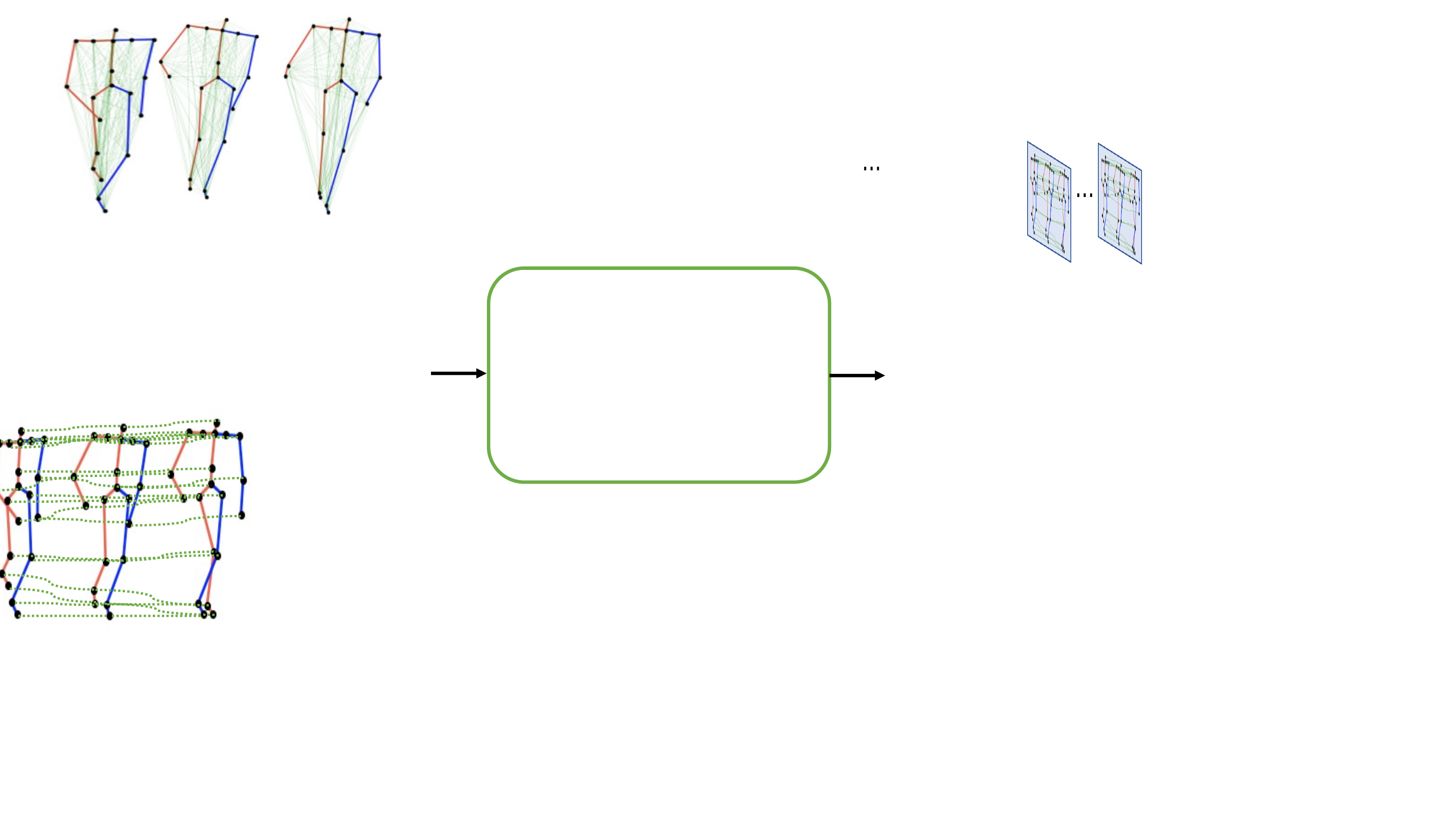}
	\caption{\footnotesize{Modelling the time series of individual joint positions}}
	\label{Fig: jp}
\end{figure}

As the first stream, we propose 3DJP-CNN that models the time series of individual joint positional information as shown in Fig. \ref{Fig: jp}. To achieve this, the 3D coordinates are modelled as channels and a 2D convolution is applied on both temporal frame and spatial joint dimensions. Modelling coordinate dimension as a channel guarantees all three-dimensional coordinates can be covered at once by a filter rather than part of them to preserve the semantics.

More specifically, the input joint positions of this stream are represented as a feature tensor $S \in R^{T \times J \times C}$, in which $T$ is the number of frames, $J$ is the number of joints, and $C=3$ represents the 3 coordinate dimensions of a joint position. Then, a 2D CNN layer is applied to obtain the extracted joint position feature $f_{jp}=Conv2d(S)$. The filter in the convolutional layer of this stream has the dimension of $F_T \times F_J \times C$, where $F_T=3$ and $F_J=1$, and the strides are both set as 1 in our experiment. With this setting, the filter covers all coordinate dimensions at once and moves along the frame and joint dimensions. This ensures that the position information of every individual joint is encoded, and the local time-series information is modeled.

\subsection{The Relative Joint Displacement Stream}
\label{sec: RJDPStream}
\begin{figure}[htbp]
	\setlength{\abovecaptionskip}{5pt} 
	\centering
	\includegraphics[scale=0.5]{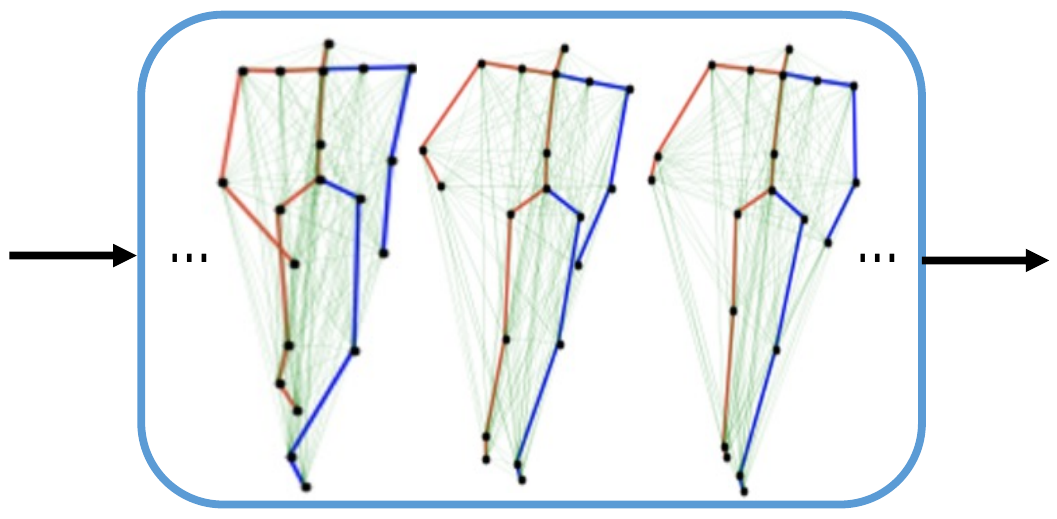}
	\caption{\footnotesize{Modelling the inter-joint correlations over time}}
	\label{Fig: 3drjdp}
\end{figure}

As the second stream, we propose 3DRJDP-CNN that models the correlations among different joint pairs over time as shown in Fig. \ref{Fig: 3drjdp}. 
Compared with the commonly used individual joint position feature \cite{SGN2020, GaitAnalysisPD2013}, the joint-pair level feature 3DRJDP is more advantageous. Because it explicitly captures the inter-joint coordination patterns and provides more information to the network preventing overfitting of learning from a small dataset. 

Concretely, given a skeleton sequence, the relative joint displacement is defined by calculating the displacement between all the joint pairs except pairs that connect to the same joint (i.e., self-loops), and it is represented as the set $\mathcal{Q}=\left\{D_{t}\left(i,j\right)\mid t=1, 2, \ldots, T ; i, j=1, 2, \ldots, J; i \neq j\right\}$, where $D_{t}\left(i,j\right)$ denotes the relative displacement of joint $i$ and $j$ at time $t$: 
\begin{equation}
	D_{t}\left(i,j\right)=\left\{\left(x_{t}^{i}-x_{t}^{j}\right),\left(y_{t}^{i}-y_{t}^{j}\right),\left(z_{t}^{i}-z_{t}^{j}\right)\right\}
\end{equation}
where $x_{t}^{i}$, $y_{t}^{i}$, and $z_{t}^{i}$ are the coordinates of joint $i$. Note that 
$D_{t}\left(i,j\right)=-D_{t}\left(j,i\right)$.

We represent the input of this stream as the feature tensor $S' \in R^{T \times D \times C}$, in which $D$ is the number of $J(J-1)$ correlations since each joint has $J-1$ correlations to all other joints. Then a 2D CNN layer is used to obtain the inter-joint feature $f_{rjdp}=Conv2d(S')$. The filter in this convolutional layer has the dimension of $F_T \times F_D \times C$, where $F_T=3$ and $F_D=J-1$, and the strides are set as 1 and $J-1$ accordingly. 
Designing the spatial filter size as $J-1$ not only enables the network to extract the inter-joint correlations of each joint but also ensures it has the same output feature size as the first stream, which facilitates the mid-layer fusion of two single streams at the feature level.

\subsection{The Two-stream Network and Feature Fusion}
\label{sec: fusion}
We present a mid-layer fusion module that takes advantage of the 3DJP and 3DRJDP streams for better prediction performance. Different types of features usually reflect different data patterns, and the fusion of them can generate a more complementary representation, facilitating better prediction performance.

There are two main advantages of mid-layer fusion architecture. Firstly, it contains much richer information on the original data \cite{feature-level-fusion2007} and does not need to train multiple classifiers compared with the late-fusion. The follow-on CNN layers further facilitate adaptively adjusting and balancing the importance of different feature sets, resulting in better prediction outcomes. Secondly, compared with an early-fusion scheme (i.e., fusing at the first layer), the CNN layers in individual schemes helps to extract useful, higher-level information from the raw features before fusion. It also ensures that the feature sizes in the two streams are compatible for fusion, since different raw feature sets are often incompatible with diverse sizes, different representation spaces of features, etc.

We fuse the outputs of individual streams in their channel dimensions as shown in our framework (Fig.~\ref{Fig: framework}). In the first stream, each value of its output tensor represents the time series of an individual joint, and in the second stream, each value represents an inter-joint correlation. The fused feature tensor $f=concat(f_{jp},f_{rjdp})$ takes the advantage from the two streams and has the dimension of $T_{2} \times J \times 2C$, it is then fed into two CNN layers for further process. After that, an adaptive Max Pooling layer is applied to extract the salient spatial-temporal information. Finally, a fully connected (FC) layer and a softmax layer are used to perform the classification. 
	
The following cross-entropy loss function is applied for the evaluation of training and testing:
\begin{equation}
	\text {L}=-\sum_{i=0}^{E-1} y_{i} \log \left(p_{i}\right)
\end{equation}
where $p=\left[p_{0}, \ldots, p_{E-1}\right]$ is a probability distribution, $p_{i}$ represents the probability that a sample belongs to class $i$. $y=\left[y_{0}, \ldots, y_{E-1}\right]$ is the one-hot representation of class labels, and $E$ is the number of classes.


\section{Experimental Results}
\label{sec:ExperimentalResults}
In this section, we first compare our proposed two-stream approach 2s-CNN against ML-based methods. Then, we discuss the performance of the two-stream model based on every single stream. Finally, the ablation study is presented.

All the results are obtained from the proposed network implemented by PyTorch and trained in an end-to-end manner with Adam optimizer. The hyper-parameters \textit{epoch}, \textit{learning rate}, and \textit{batch size} are set as 80, 0.003 and 57 respectively. Our evaluation is done by 5-fold cross-validation. The averaged outcome for all cross-validations is presented as the final result.

\begin{table}
\begin{center}
\begin{minipage}{\linewidth}
\caption{Comparisons with other methods in accuracy (\%)}\label{comparewithml}%
\begin{tabular}{@{}cccccc@{}}
\toprule
Method&Healthy&\tabincell{c}{Joint\\ Problem}&\tabincell{c}{Muscle\\ Weakness}&\tabincell{c}{Neurological\\ Defect}&Average\\
\midrule
	Naive Bayes \cite{bayes}&\textbf{100.00}&75.00&94.44&53.85&82.22\\
	Random Forest \cite{decisiontree-randfor}&\textbf{100.00}&75.0&83.33&\textbf{92.31}&88.89\\
	Decision Tree \cite{Worasak2018,decisiontree-randfor}&\textbf{100.00}&75.00&77.78&84.62&84.44\\
	SVM (RBF) \cite{Worasak2018}&\textbf{100.00}&\textbf{100.00}&0.00&23.08&37.78\\
	SVM (Linear) \cite{Worasak2018}&\textbf{100.00}&75.00&83.33&\textbf{92.31}&88.89\\
	SVM (Polynomial) \cite{Worasak2018}&\textbf{100.00}&75.00&83.33&\textbf{92.31}&88.89\\ 
	FCNet \cite{KevinIEEEAccess}&\textbf{100.00}&75.00&77.78&\textbf{92.31}&86.67\\ \hline
	3DJP-CNN&90.00&75.00&94.44&84.62&88.89\\
	3DRJDP-CNN&\textbf{100.00}&75.00&94.44&84.62&91.11\\
	2s-CNN&\textbf{100.00}&75.00&\textbf{100.00}&\textbf{92.31}&\textbf{95.56}\\ 
\botrule
\end{tabular}
\end{minipage}
\end{center}
\end{table}
	
\subsection{Comparisons with the Other Methods}
We compare our method with the start-of-the-art SVM approach with different kernels \cite{Worasak2018} and some other ML-based methods, including Decision Tree \cite{Worasak2018,decisiontree-randfor}, Bayes \cite{bayes}, and Random Forest \cite{decisiontree-randfor}. We also compare with the fully connected deep network architecture FCNet that was designed in \cite{KevinIEEEAccess} for cerebral palsy prediction. To make a fair comparison, both 3DJP and 3DRJDP features are used in all these methods, and hyperparameters are turned to an optimal configuration in our experiment.
	
It can be seen in Table \ref{comparewithml} that our proposed 2s-CNN achieves the best average accuracy of 95.56\% on all classes, and the 3DRJDP-CNN outperforms 3DJP-CNN, FCNet, and ML-based methods. This demonstrates the power of convolutional neural networks and the advantage of fusing 3DJP and 3DRJDP features. 

\begin{figure*}[htbp]
	\centering
	\subfigure[3DJP-CNN]{
		\begin{minipage}{0.45\linewidth}
			\includegraphics[scale=0.3, left]{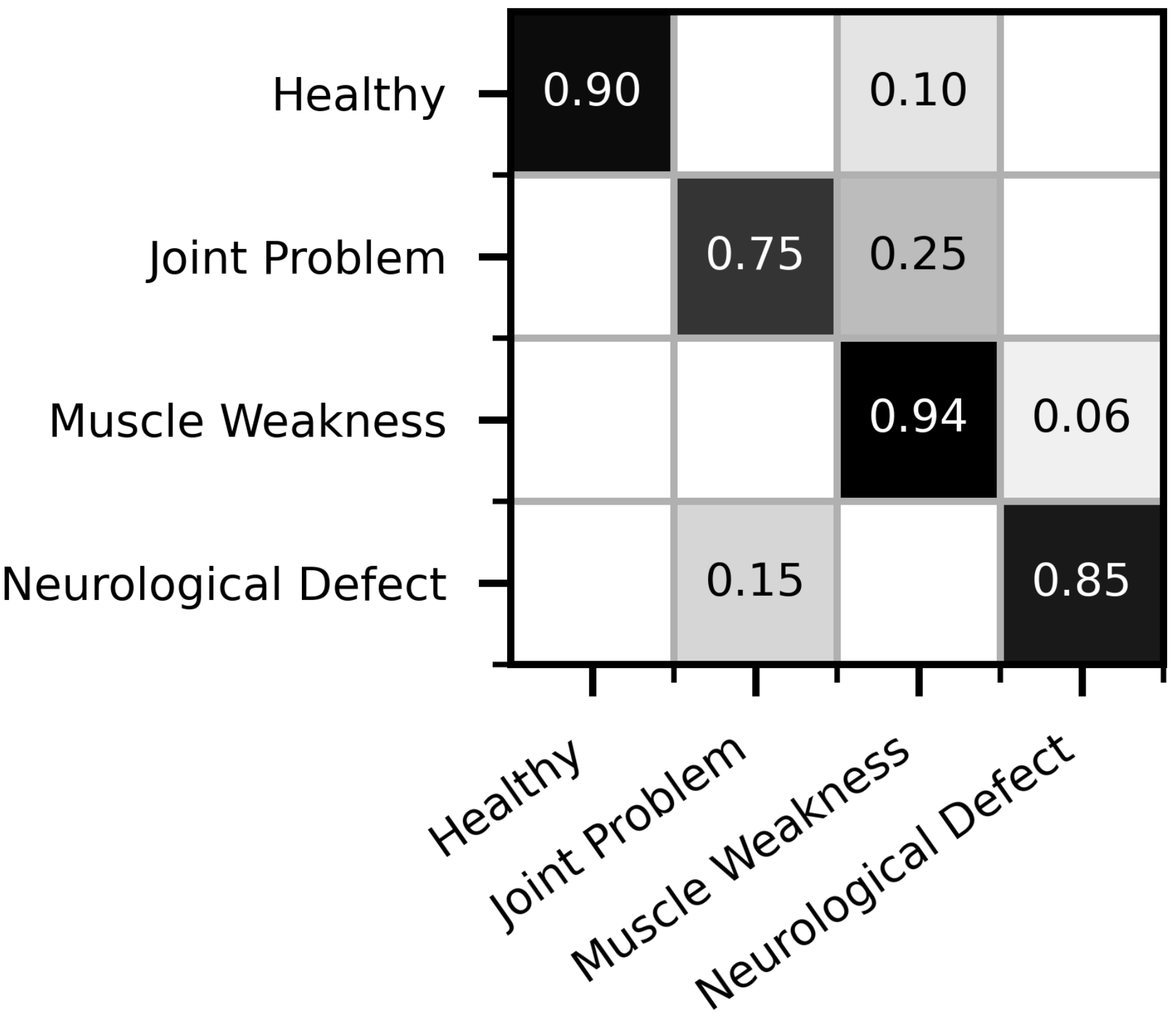}
		\end{minipage}
	}
    \quad
	\subfigure[3DRJDP-CNN]{
		\begin{minipage}{0.45\linewidth}
			\includegraphics[scale=0.3,left]{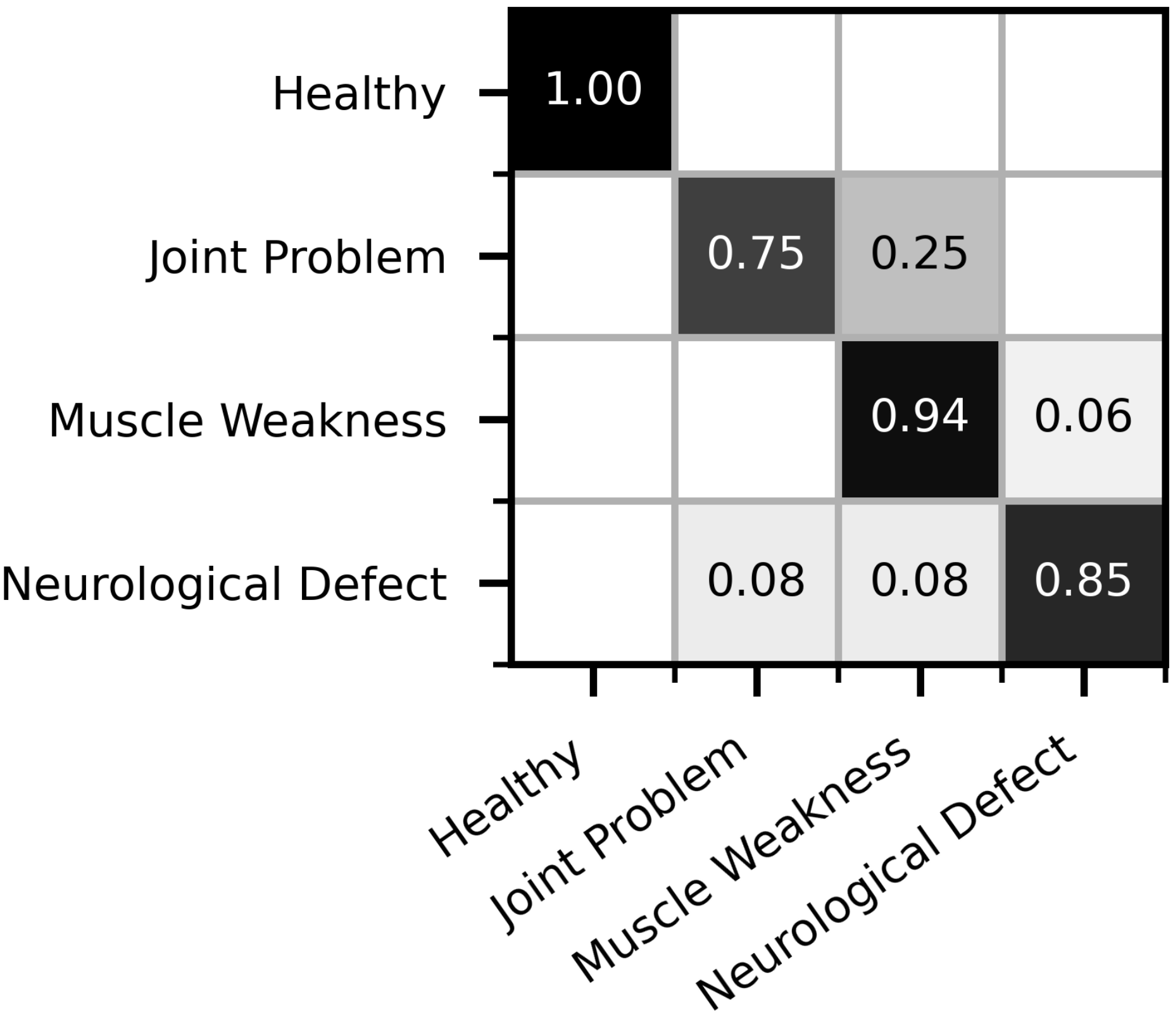} 
		\end{minipage}
	}
    \quad
	\subfigure[2s-CNN]{ 
		\begin{minipage}{0.5\linewidth}
			\centering
			\includegraphics[scale=0.3]{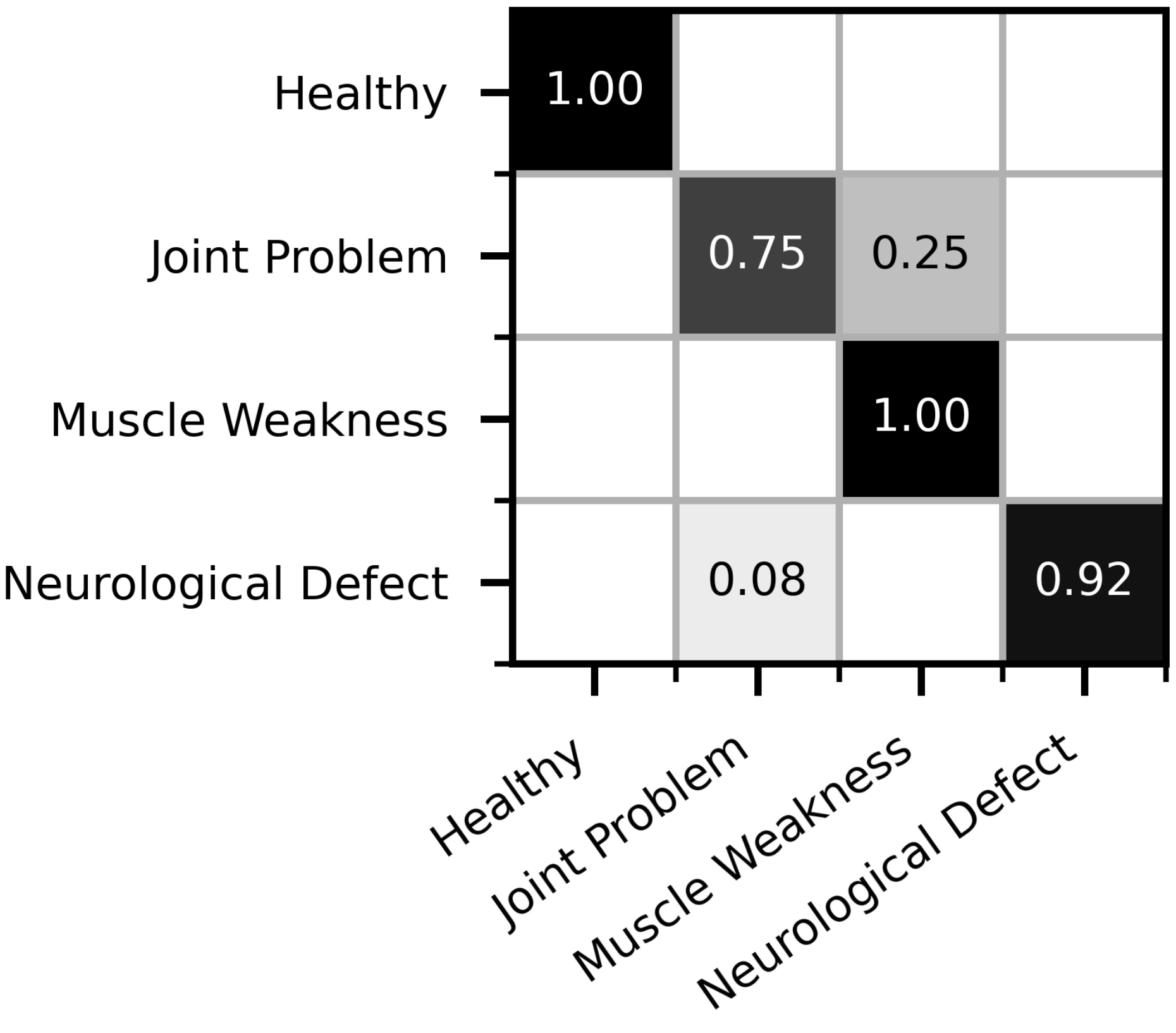} 
		\end{minipage}
	}
	\caption{Confusion matrices of single-stream and the two-stream networks} 
	\label{cm} 
\end{figure*}

\subsection{Comparisons with Baselines}
We build two single-stream networks as baselines. The classification accuracy of our two-stream network and baselines is shown in Table \ref{comparewithml} (the last three lines). We observe that 2s-CNN outperforms baselines and achieves a significant accuracy improvement of 4.4\%. This demonstrates the two feature sets are complementary to each other under our fusion design. In addition, 3DRJDP-CNN performs on par with or better than 3DJP-CNN in all classes. We believe that this is because 3DRJDP carries explicit inter-joint correlations, which makes it more powerful in distinguishing healthy and unhealthy gaits.

\begin{figure*}[htbp]
	\centering
	\subfigure[ROC Curve for 3DJP-CNN model]{
		\begin{minipage}{0.44\linewidth}
			\includegraphics[scale=0.35, left]{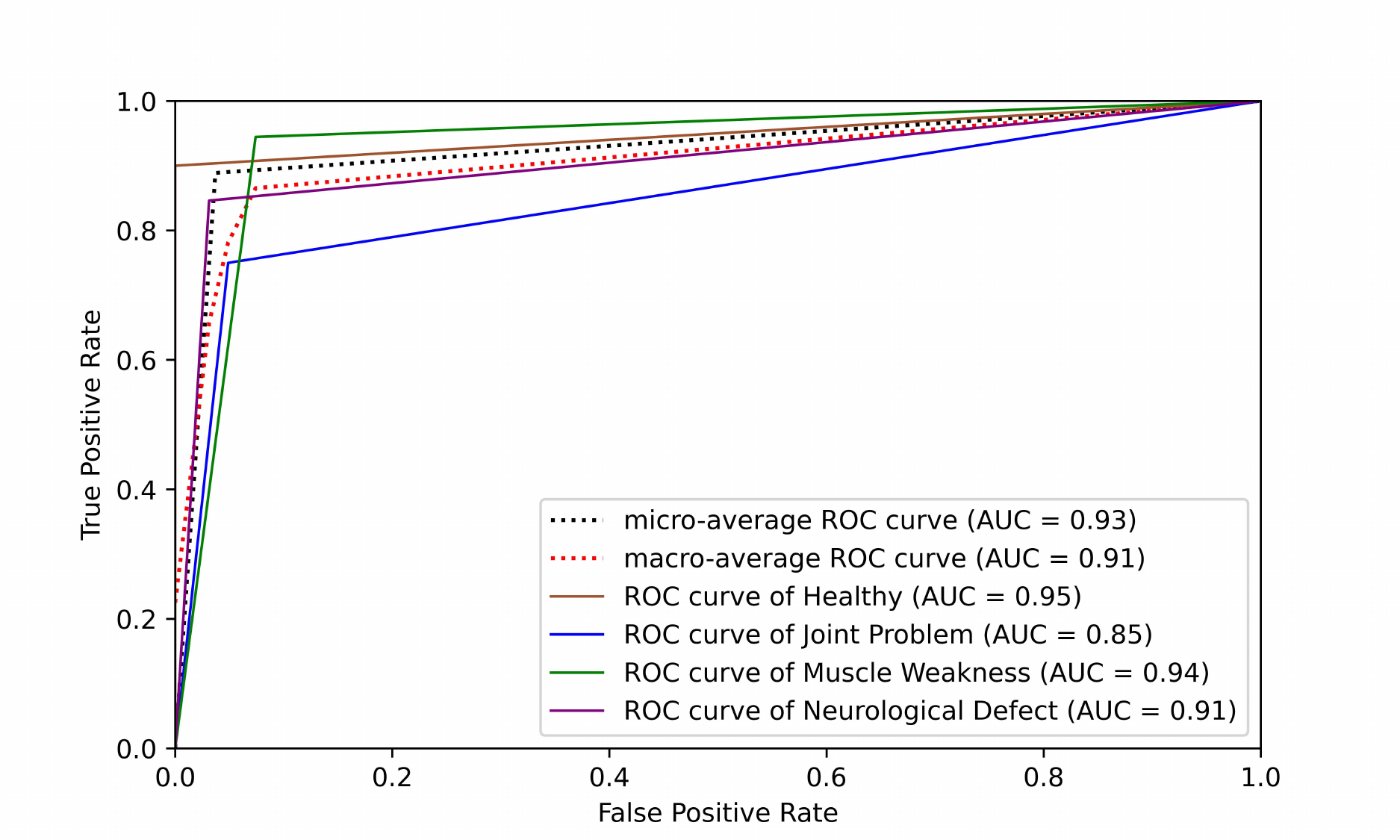}
		\end{minipage}
	}
    \quad
	\subfigure[ROC Curve for 3DRJDP-CNN model]{
		\begin{minipage}{0.4\linewidth}
			\includegraphics[scale=0.35,left]{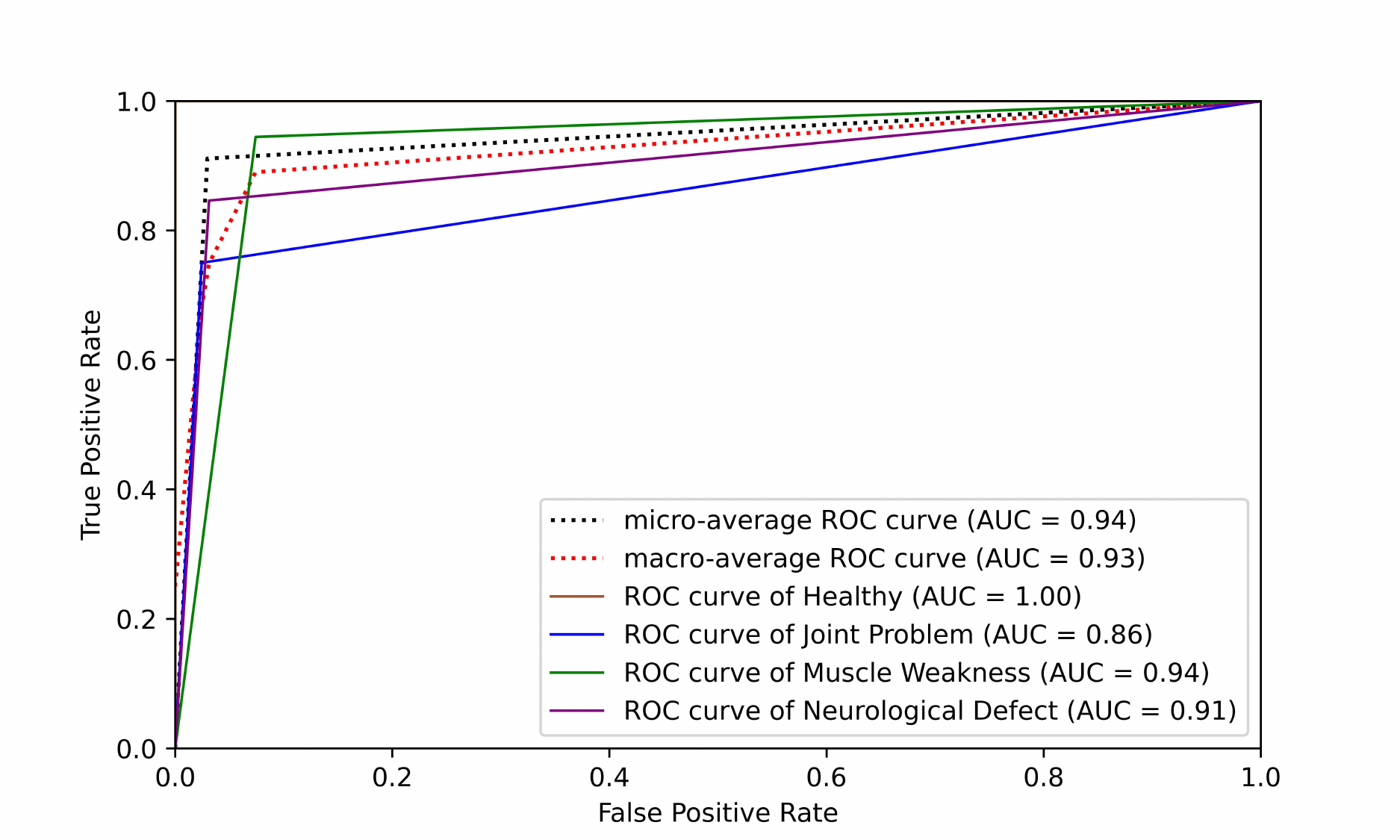} 
		\end{minipage}
	}
    \quad
	\subfigure[Roc Curve for 2s-CNN model.]{ 
		\begin{minipage}{0.5\linewidth}
			\centering
			\includegraphics[scale=0.35]{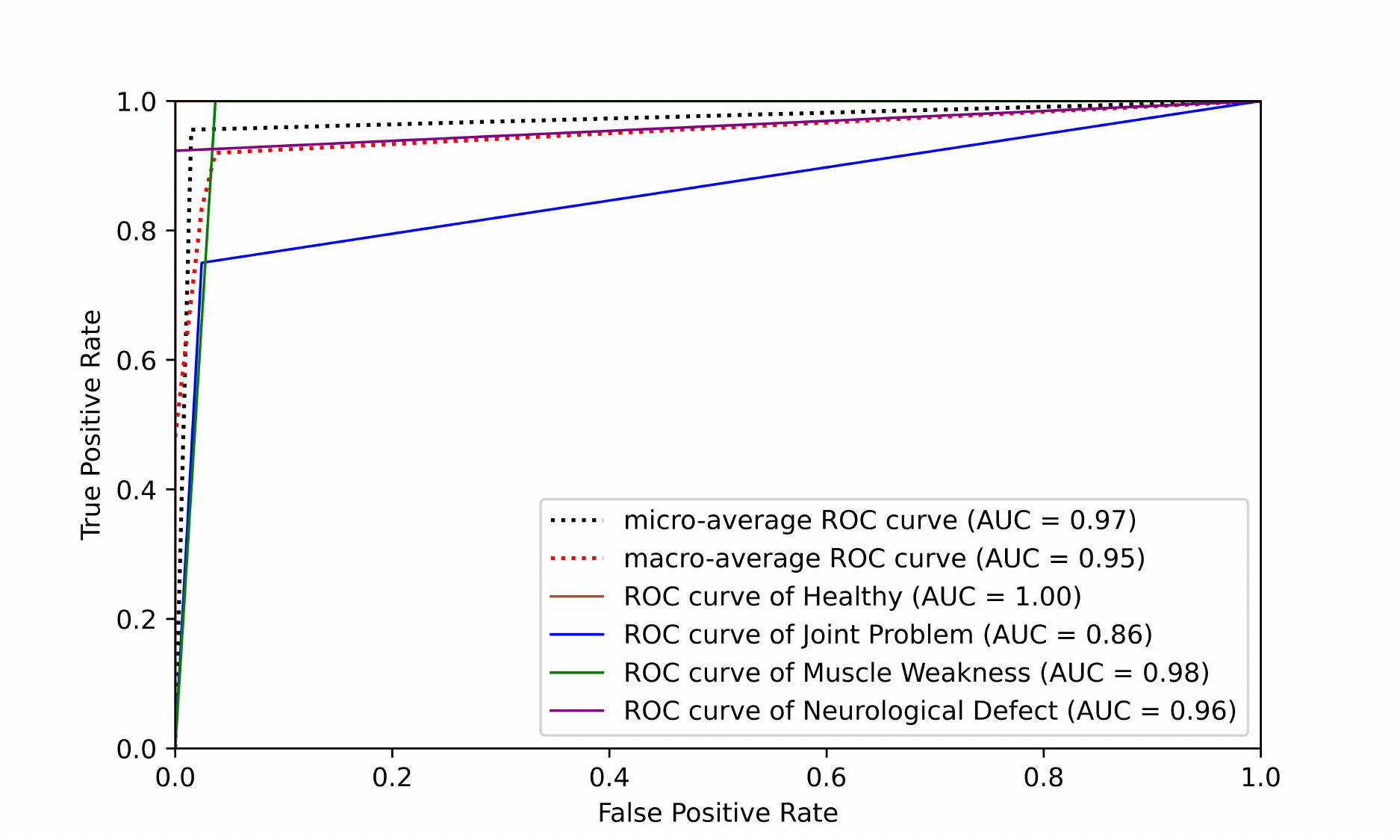} 
		\end{minipage}
	}
	\caption{Receiver operating characteristic curves for multi-class disorder classification} 
	\label{roc} 
\end{figure*}
	
To investigate the contributions of single networks and the improvement of the fusion module for different gait disorders and the healthy classes, Fig. \ref{cm} presents the confusion matrices. We observe that 3DJP-CNN fails to differentiate between healthy and unhealthy classes, and 3DRJDP suffers for distinguishing three unhealthy classes. We argue that this is because 3DJP-CNN does not contain explicit inter-joint correlations. More importantly, 2s-CNN takes advantage of 3DJP-CNN and 3DRJDP-CNN by only struggling in two unhealthy classes, i.e., joint problem and neurological defects. This indicates that the fusion module is having a positive impact on the overall classification performance. 

\begin{table}
	\centering
	\caption{Comparison of results with baselines on precision, recall, f1-measure, and AUC}
	\label{metrics}
	\setlength{\tabcolsep}{4.5pt}
	\begin{tabular}{ccccccc}\hline
	Metric & Network & Healthy & \tabincell{c}{Joint\\Problem} & \tabincell{c}{Muscle\\Weakness} & \tabincell{c}{Neurological\\ Defect} & Average\\ \hline
	
	\multirow{3}{*}{Precision} & 3DJP-CNN & \textbf{1.00} & 0.60 & 0.89 & 0.92 & 0.85\\ 
	& 3DRJDP-CNN & \textbf{1.00} & \textbf{0.75} & 0.89 & 0.92 & 0.89\\
	& 2s-CNN & \textbf{1.00} & \textbf{0.75} & \textbf{0.95} & \textbf{1.00} & \textbf{0.92}\\
	\hline
	\multirow{3}{*}{Recall} & 3DJP-CNN & 0.90 & \textbf{0.75} & 0.94 & 0.85 & 0.86\\ 
	& 3DRJDP-CNN & \textbf{1.00} & \textbf{0.75} & 0.94 & 0.85 & 0.89\\
	& 2s-CNN & \textbf{1.00} & \textbf{0.75} & \textbf{1.00} & \textbf{0.92} & \textbf{0.92}\\
	\hline
	\multirow{3}{*}{F1-Measure} & 3DJP-CNN & 0.95 & 0.67 & 0.92 & 0.88 & 0.85\\ 
	& 3DRJDP-CNN & \textbf{1.00} & \textbf{0.75} & 0.92 & 0.88 & 0.89\\
	& 2s-CNN & \textbf{1.00} & \textbf{0.75} & \textbf{0.97} & \textbf{0.96} & \textbf{0.92}\\	
	\hline
	\multirow{3}{*}{AUC} & 3DJP-CNN & 0.95 & 0.85 & 0.94 & 0.91 & 0.91\\ 
	& 3DRJDP-CNN & \textbf{1.00} & \textbf{0.86} & 0.94 & 0.91 & 0.93\\
	& 2s-CNN & \textbf{1.00} & \textbf{0.86} & \textbf{0.98} & \textbf{0.96} & \textbf{0.95}\\ \hline
	\end{tabular}
\end{table}

\begin{figure*}[htbp]
	\centering 
	\subfigure{
		\begin{minipage}{8cm}
			\centering 
			\includegraphics[scale=0.48]{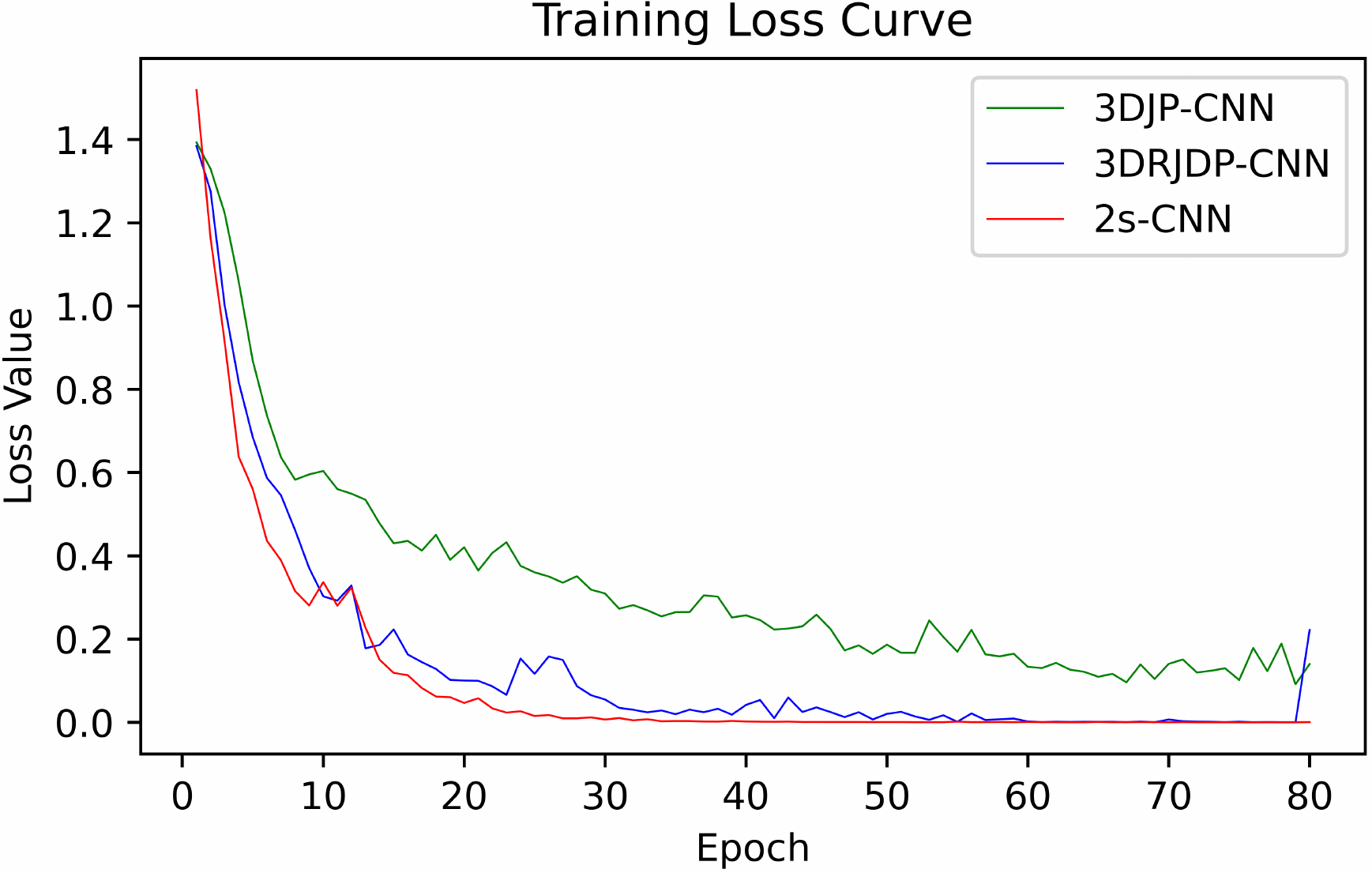} 
		\end{minipage}
	}
	\quad
	\subfigure{ 
		\begin{minipage}{8cm}
			\centering
			\includegraphics[scale=0.48]{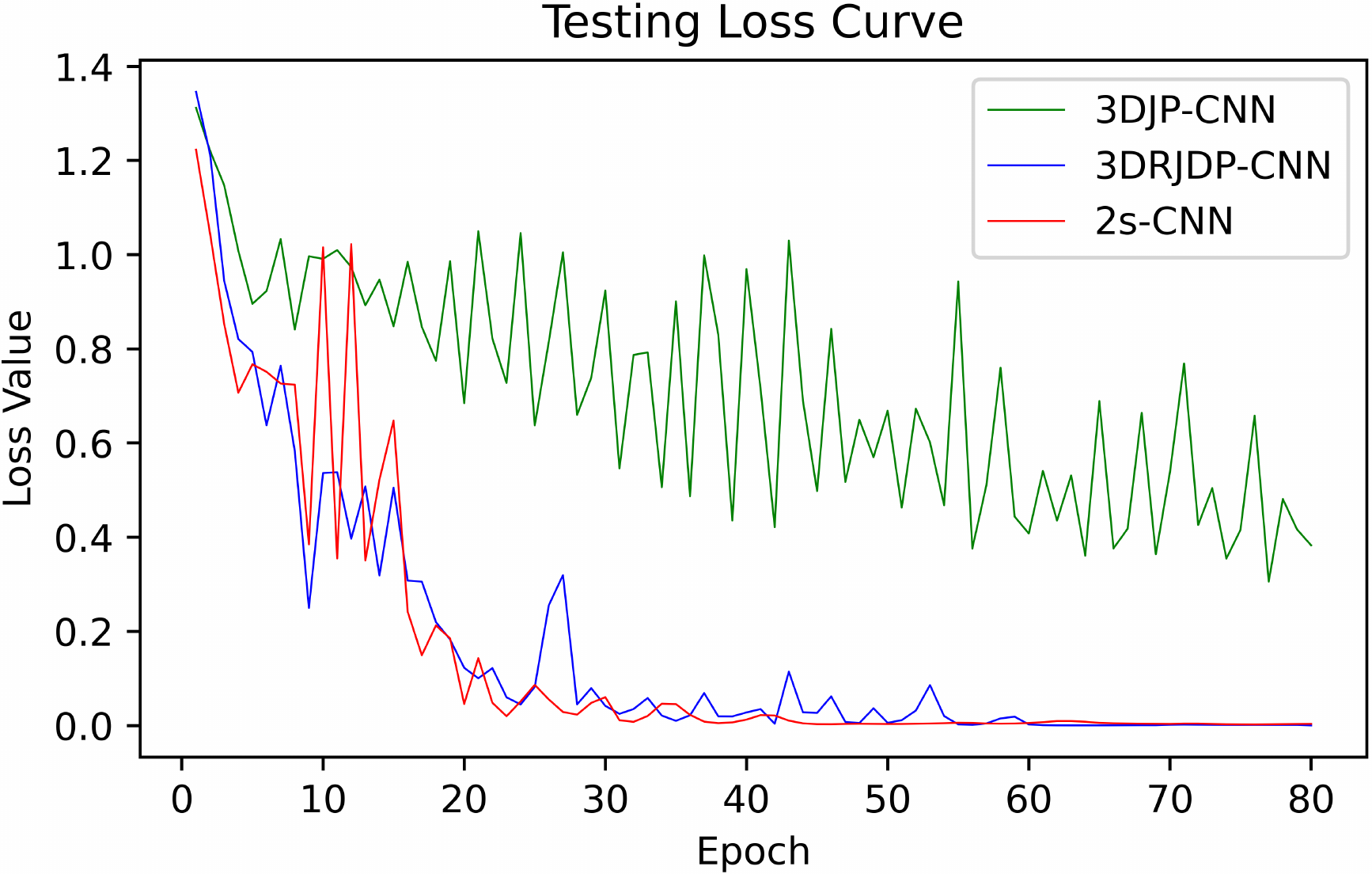} 
		\end{minipage}
	}
	\caption{Training and testing loss curves of single-stream and two-stream networks} 
	\label{loss} 
\end{figure*}

To make a comprehensive comparison between the proposed two-stream network and baselines, besides accuracy, precision, recall, f1-measure, receiver operating characteristic (ROC) curves, and AUC (area under the receiver operating characteristic) are also reported in Table \ref{metrics} and Fig. \ref{roc}. The results show that our proposed 2s-CNN model achieves consistent superior performance on these measure metrics in both individual class and average evaluations.

To further demonstrate the robustness of the proposed methods, their training and testing loss curves are presented in Fig. \ref{loss}. 
It is can be seen that 2s-CNN generates more consistent and robust curves than 3DJP-CNN and 3DRJDP-CNN, and 3DRJDP-CNN performs better than 3DJP-CNN. This aligns with the results of the abovementioned evaluation results.

\begin{figure}[htbp]
	\centering 
	\subfigure[No CNN layers]{
		\begin{minipage}{3.5cm}
			\centering 
			\includegraphics[scale=0.35]{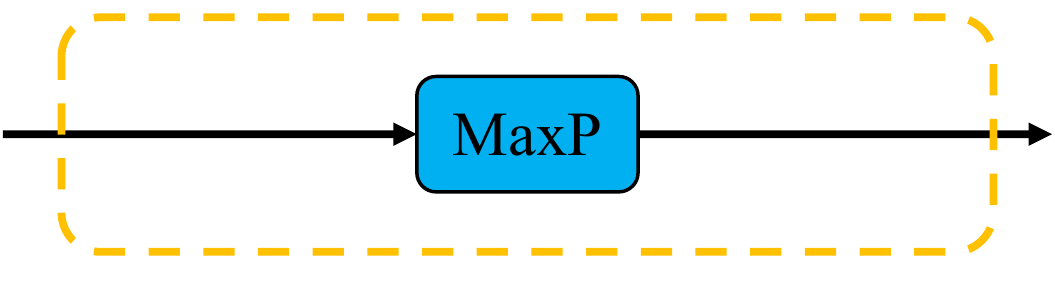} 
		\end{minipage}
	}
	\quad
	\subfigure[No MaxP layer]{ 
		\begin{minipage}{3.5cm}
			\centering
			\includegraphics[scale=0.35]{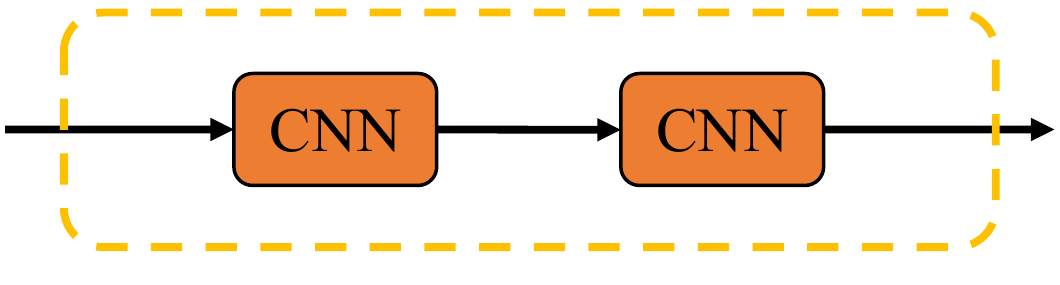} 
		\end{minipage}
	}
	\quad
	\subfigure[Single CNN layer]{ 
	\begin{minipage}{3.5cm}
		\centering
		\includegraphics[scale=0.35]{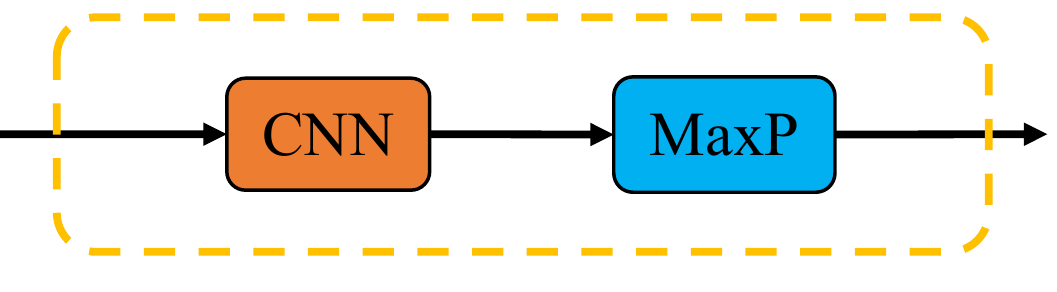} 
	\end{minipage}
	}
    \quad
    \subfigure[Ours (2 CNN + MaxP)]{ 
	\begin{minipage}{3.5cm}
		\centering
		\includegraphics[scale=0.35]{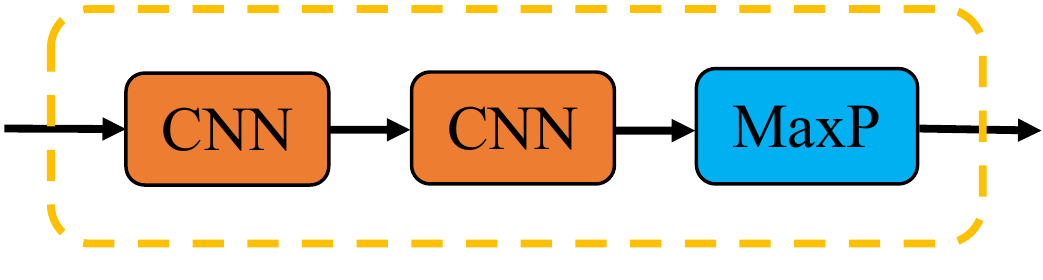} 
	\end{minipage}
    }
	\caption{Proposed fusion network architecture with different CNN and MaxP combinations}
	\label{subpart} 
\end{figure}

\begin{table}[htbp]
	\centering
	\caption{The number of parameters, averaged per-sample test time, and accuracy of proposed fusion network architecture with different CNN and Max Pooling combinations}
	\label{compsub}
	\setlength{\tabcolsep}{3pt}
	\begin{tabular}{{l}c c c c} \hline
		\hspace{0.4in} Method & \#Params & Test Time & Accuracy\\ \hline
		No CNN layers (No-CNN) & 86288 & 44.53ms & 91.11\\ 
		No Max Pooling layer (No-MaxP) & 238864 & \textbf{44.36ms} & 93.33\\ 
		Single CNN layer (SinCNN) & \textbf{86032} & 46.94ms & 93.33\\ \hline
		Ours (2 CNN + MaxP) & 233488 & 47.29ms & \textbf{95.56}\\ \hline
	\end{tabular}
\end{table}

\subsection{Ablation Study}
To validate our network architecture, we compare our proposed fusion network architecture under different CNN and Max Pooling (MaxP) combinations as shown in Fig. \ref{subpart}. The number of parameters, averaged per-sample test time, and accuracy are reported in Table \ref{compsub}. We observe that the system only can achieve an accuracy of 91.11\% without any CNN layers (No-CNN).
The two-CNN architecture (No-MaxP) has no improvement compared with a single CNN layer (SinCNN), but it improves with the help of a Max Pooling layer (Ours). This is because two CNN layers have digested the majority of the features, and Max Pooling could further select the discriminative information. Notice that although our two-CNN architecture contains more parameters, the test time only demonstrates a slight increase.

\begin{figure*}[htbp]
		\begin{minipage}{0.3\linewidth}
			\centering
			\includegraphics[scale=0.8]{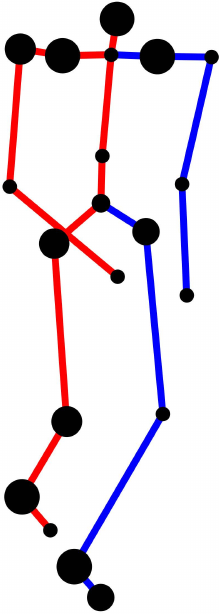} 
		\end{minipage}
		\begin{minipage}{0.3\linewidth}
			\centering
			\includegraphics[scale=0.8]{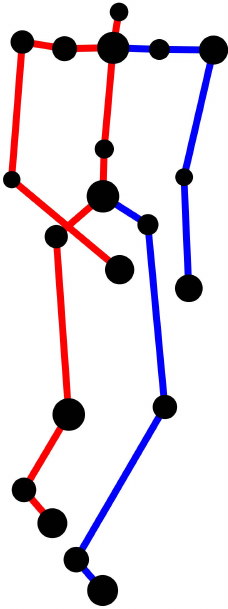} 
		\end{minipage}
		\begin{minipage}{0.35\linewidth}
			\centering
			\includegraphics[scale=0.8]{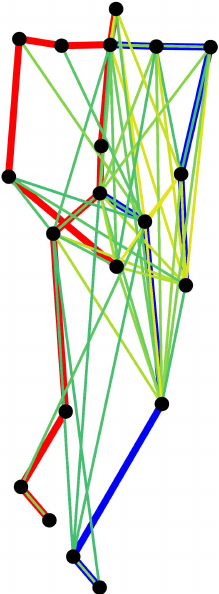} 
		\end{minipage}
	\subfigure[Single joint importance]{ 
		\begin{minipage}{0.3\linewidth}
			\centering
			\includegraphics[scale=0.8]{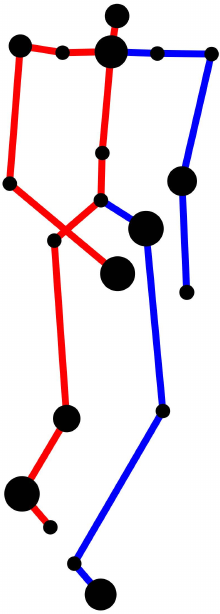} 
		\end{minipage}
	}
	\subfigure[Relative joint displacements attention values aggregated to each joint]{ 
		\begin{minipage}{0.3\linewidth}
			\centering
			\includegraphics[scale=0.8]{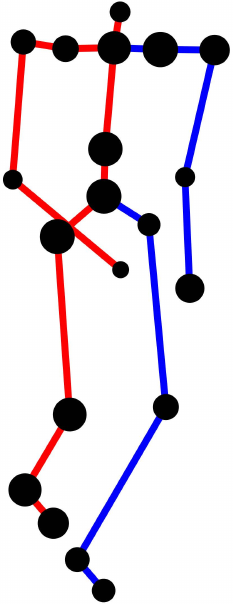} 
		\end{minipage}
	}
	\subfigure[Top 50 important relative joint displacements]{ 
		\begin{minipage}{0.3\linewidth}
			\centering
			\includegraphics[scale=0.8]{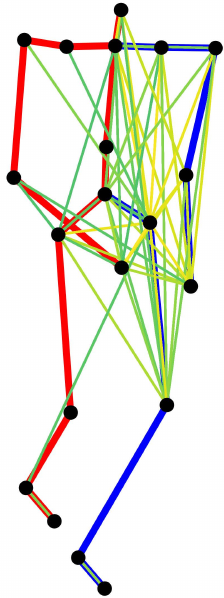} 
		\end{minipage}
	}
	\caption{Visualization of the importance of different joints and relative joint displacements with a healthy sample (upper row) and a muscle weakness sample (lower row). The larger size of black joints represent higher importance. The relative joint displacements attention values are aggregated to each joint to show which joints have more interactions with other joints. The importance of relative joint displacements is visualized from a yellow to emerald green scale, with the yellow color representing higher importance.}
	\label{attention} 
\end{figure*}

We further visualize the importance of joints and relative joint displacements across validations from 3DJP-CNN (with joint position features) and 3DRJDP-CNN (with relative joint displacement features) streams respectively using channel attention \cite{senet2018} as shown in Fig. \ref{attention}. For single joint importance, we can see that the two streams mostly demonstrate different importance on the same joints, indicating that they focus on different aspects of a joint. Notice that they both have higher importance on the foot joints, this may be because the subject's balance is impaired by walking issues, e.g., the foot/ankle proprioceptive input is decreased \cite{gaitbalance}. More importantly, the 3DRDJP-CNN stream assists in identifying which joint pairs are highly involved in the body movements, i.e., the left upper body joints tend to have more interactions. This could be because the subject tries to avoid using the right body parts due to pain, resulting in imbalanced gaits \cite{painimbalance}. In addition, the body parts' importance of the two samples shares similar patterns, indicating our model is consistent across classes. This kind of human-understanding visualization can effectively support clinicians for in-depth analysis, e.g., rapid lesion locating.

\section{Conclusion}
\label{sec:Conclusion}
In this paper, we have proposed a 2s-CNN framework that explicitly takes both individual joint features and inter-joint features as input for musculoskeletal and neurological disorders prediction. Our proposed mid-layer fusion module adaptively merges individual joint 3DJP and inter-joint 3DRJDP features into the network to jointly learn and update with the model, relieving the system from the need of discovering more complicated features from small data. The experimental results have shown that the inter-joint 3DRJDP features demonstrated more effectiveness for different disorders classification, which aligned with the intuition that movement is generated by body parts' coordination. The method \textit{mixup} \cite{mixup} was used to deal with the data bias problem, resulting in a more robust system. We demonstrated the effectiveness of the mid-layer fusion of fusing the two sets of features. 
Compared with ML-based methods and the fully connected deep network, our proposed model outperforms them with a better average prediction accuracy of 95.56\%. The accuracy of every individual class is also reported for the first time.

Interpreting DL models is critical in the medical area because it can provide us with more insights into these advanced automatic tools, thus gaining the trust of clinicians and patients. Our interpretable visualization of spatial attention facilitates a user to focus the analysis on the body parts with high attention. For future study, we intend to interpret the proposed framework from both spatial and temporal domains for frame-to-frame interpretation, to generate a more solid automatic musculoskeletal and neurological disorders prediction system. As the current dataset is relatively small, we will consider enlarging it in future works, e.g., increasing the number of subjects and the video length including walking cycles, such that other important factors (e.g., step-to-step variability) could be better included for a more robust diagnostic system.

\bmhead{Compliance with Ethical Standards}
The authors have no relevant financial or non-financial interests to disclose. The motion data used in this research is shared by \cite{Worasak2018} as part of the contributions in their published work and they obtained the ethics approval when capturing the data. 

\bmhead{Acknowledgments}

This work was supported in part by the Royal Society (Ref: IES\textbackslash R1\textbackslash 191147).



\bibliography{main.bbl}


\end{document}